\documentclass{article}

\PassOptionsToPackage{numbers}{natbib}

\usepackage[preprint]{neurips_data_2023}

\usepackage[utf8]{inputenc} 
\usepackage[T1]{fontenc}    
\usepackage{hyperref}       
\hypersetup{
    colorlinks,
    linkcolor={red!85!black},
    citecolor={blue!70!cyan},
    urlcolor={blue!90!black}
}
\usepackage{url}            
\usepackage{booktabs}       
\usepackage{authblk}        
\usepackage{amsfonts}       
\usepackage{nicefrac}       
\usepackage{microtype}      
\usepackage{xcolor}         

\usepackage{colortbl}
\usepackage{multirow}

\usepackage{wrapfig}

\usepackage{pifont}
\usepackage{amsmath}

\newcommand{\resolved}[1]{}

\usepackage[pdftex]{graphicx}

\usepackage{array}
\makeatletter
\def\hlinew#1{%
  \noalign{\ifnum0=`}\fi\hrule \@height #1 \futurelet
   \reserved@a\@xhline}
\makeatother

\newcommand{\printfnsymbol}[1]{%
  \textsuperscript{\@fnsymbol{#1}}%
}

\title{GPT-ImgEval: A Comprehensive Benchmark for Diagnosing GPT4o in Image Generation}

\author{
  \textbf{Zhiyuan Yan}\textsuperscript{1,3*},
  \textbf{Junyan Ye}\textsuperscript{2,4*},
  \textbf{Weijia Li}\textsuperscript{2$^\dagger$},
  \textbf{Zilong Huang}\textsuperscript{2},
  \textbf{Shenghai Yuan}\textsuperscript{1,3}, \\
  \textbf{Xiangyang He}\textsuperscript{6}, 
  \textbf{Kaiqing Lin}\textsuperscript{5},
  \textbf{Jun He}\textsuperscript{2},
  \textbf{Conghui He}\textsuperscript{4},
  \textbf{Li Yuan}\textsuperscript{1$^\dagger$}
}

\affil{
  {\tt 
  $*$ Equal Contributors, $\dagger$ Corresponding Authors
  }\\
  \textsuperscript{1}Peking University, Shenzhen Graduate School,
  \textsuperscript{2}Sun Yat-sen University, \\
  \textsuperscript{3}Rabbitpre AI,
  \textsuperscript{4}Shanghai AI Laboratory,
  \textsuperscript{5}Shenzhen University \\
  \textsuperscript{6}The Hong Kong University of Science and Technology (Guangzhou), \\
  {\tt 
  \{zhiyuanyan@stu.,yuanli-ece@\}pku.edu.cn, \\ yejy53@mail2.sysu.edu.cn, liweij29@mail.sysu.edu.cn
  }
}

\begin{document}

\maketitle

\begin{abstract}
The recent breakthroughs in OpenAI's GPT4o model have demonstrated surprisingly good capabilities in image generation and editing, resulting in significant excitement in the community.
This technical report presents the first-look evaluation benchmark (named \textbf{GPT-ImgEval}), quantitatively and qualitatively diagnosing GPT-4o's performance across three critical dimensions: (1) generation quality (assessed through the GenEval dataset), (2) editing proficiency (measured via the Reason-Edit dataset), and (3) world knowledge-informed semantic synthesis (evaluated using the WISE dataset). Across all three tasks, GPT-4o demonstrates strong performance, significantly surpassing existing methods in both image generation control and output quality, while also showcasing exceptional knowledge reasoning capabilities. 
Furthermore, based on the GPT-4o's generated data, we propose a classification-model-based approach to investigate the underlying architecture of GPT-4o, where our empirical results suggest the model consists of an auto-regressive (AR) combined with a diffusion-based head for image decoding, rather than the VAR-like architectures. We also provide a complete speculation on GPT-4o's overall architecture.
In addition, we conduct a series of analyses to identify and visualize GPT-4o's specific limitations and the synthetic artifacts commonly observed in its image generation.
We also present a comparative study of multi-round image editing between GPT-4o and Gemini 2.0 Flash, and discuss the safety implications of GPT-4o's outputs, particularly their detectability by existing image forensic models.
We hope that our work can offer valuable insight and provide a reliable benchmark to guide future research, foster reproducibility, and accelerate innovation in the field of image generation and beyond. 
The codes and datasets used for evaluating GPT-4o can be found at \url{https://github.com/PicoTrex/GPT-ImgEval}.
\end{abstract}

\section{Introduction}
\label{sec:intro}

Recent advances in multimodal large language models (MLLMs) have brought remarkable progress in unified vision-language understanding and generation~\cite{wang2024emu3, wu2410janus, ma2024janusflow, janus-pro, unitok, show-o, metamorph, llamafusion, transfusion, fan2025unified, song2025dualtoken, chen2025semhitok, nova, seed-x}. 
Among these, OpenAI’s newly released GPT-4o\footnote{\url{https://openai.com/index/gpt-4o-system-card/}} (where the "o" stands for "omni") has attracted widespread attention due to its surprising proficiency in image generation, editing, and vision-language reasoning—all within a single unified architecture. 
With the increasingly growing use of GPT4o in real-world applications such as digital content creation, and interactive assistants, \textit{it becomes increasingly critical and necessary to systematically evaluate its image generation capabilities, weaknesses, failure cases, and other related problems in generative settings.}

\begin{figure}[!t]
    \centering
    \includegraphics[width=\linewidth]{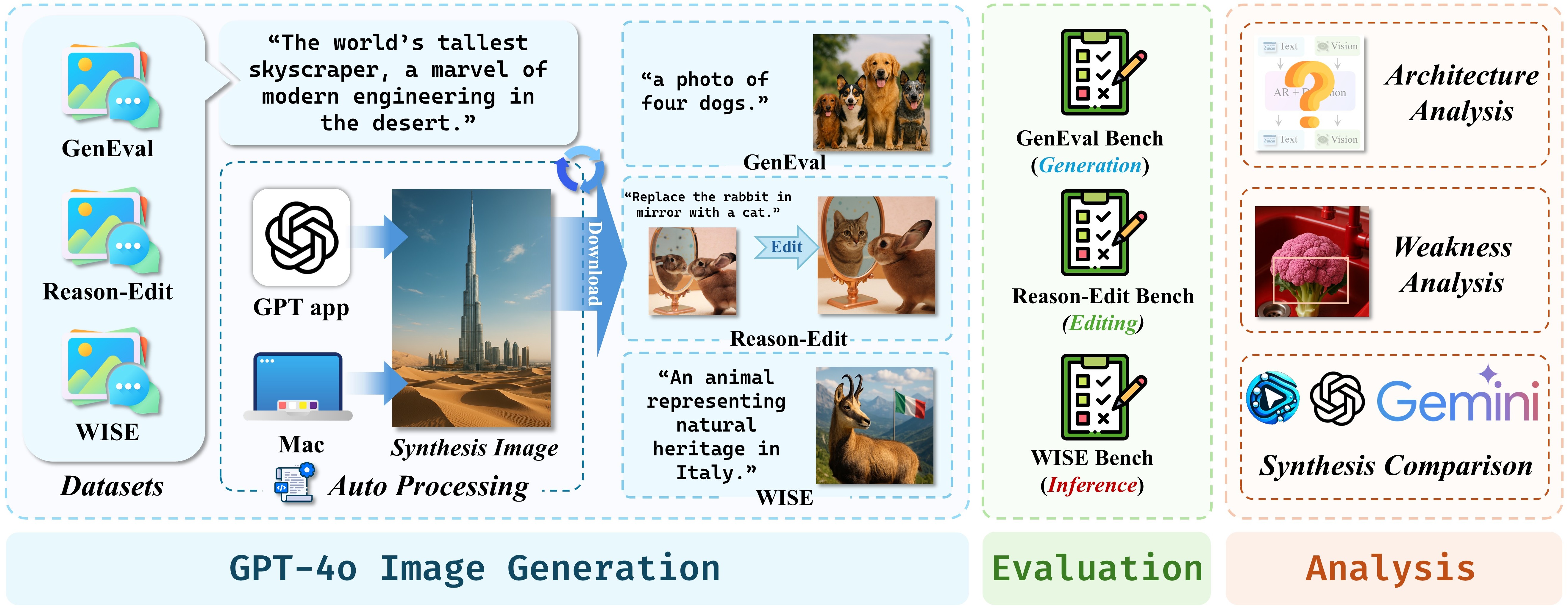}
    \caption{The overall workflow of our \textbf{GPT-ImgEval}, consisting the GPT-4o Image generation, Evaluation, and Analysis.}
    \label{fig:pipeline}
    \vspace{-3mm}
\end{figure}

To fill this gap, we present \textbf{GPT-ImgEval}, \textit{the \textbf{first comprehensive benchmark} designed to evaluate GPT-4o’s capabilities in image generation}. The overall workflow of our benchmark is illustrated in Figure~\ref{fig:pipeline}. We focus on three core image-generation tasks: text-to-image generation \cite{flux2024, Fluid, opensora, generative_pretraining, ye2024loki,NPP}, instruction-based image editing \cite{tian2025mige, chen2025multimodal, ultraedit, instructpix2pix, magicbrush, easycontrol, instantid, uniportrait}, and world knowledge-informed semantic synthesis \cite{T2i-compbench, phybench, Commonsense-T2I}.
For each task, we adopt either existing or purpose-built benchmarks: GenEval~\cite{ghosh2023geneval} for text-to-image generation, Reason-Edit~\cite{huang2024smartedit} for instruction-guided editing, and WISE~\cite{niu2025wise} for evaluating knowledge-grounded semantic understanding. \textbf{Across all three tasks, GPT-4o demonstrates strong performance}, showcasing accurate compositional reasoning, fine-grained attribute control, and a nuanced understanding of real-world context.

\begin{figure}[t]
    \centering
    \includegraphics[width=\linewidth]{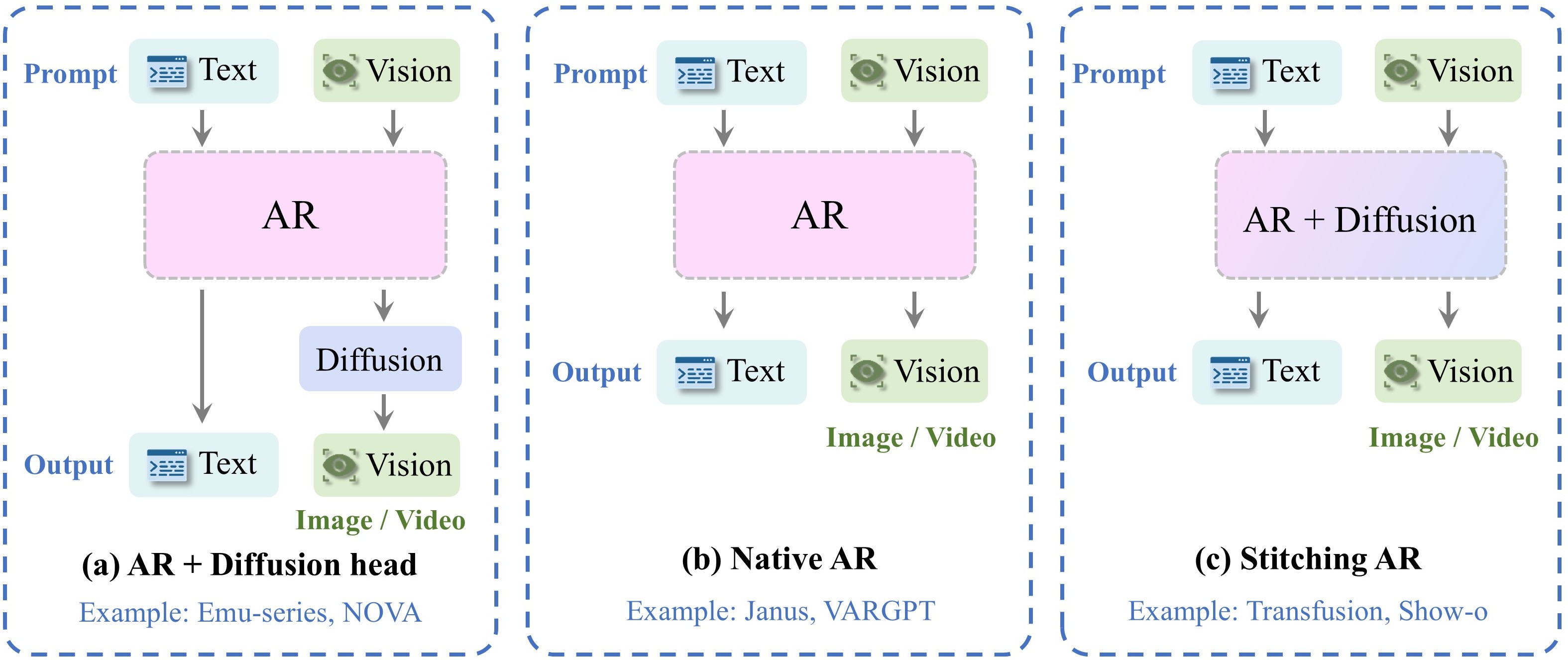}
    \caption{Commonly used pipelines for unified image generation and understanding, and potential decoder architectures of GPT4o's image generation choice. The complete speculation of architectures can be seen in Figure~\ref{fig:complete}.}
    \label{fig:archi}
    \vspace{-2mm}
\end{figure}

Beyond benchmark evaluations, we conduct deeper analyses to \textbf{uncover GPT-4o's potential architectural choices}. Specifically, we first explore whether GPT-4o relies on a diffusion-based or autoregressive decoder head (see Figure~\ref{fig:archi}). To this end, we propose a model-based classification method, where a standard binary classifier is trained to distinguish between images generated by the two paradigms, and then applied to GPT-4o's outputs. Interestingly, the classifier consistently classifies GPT-4o's images as diffusion-based, providing empirical evidence that \textbf{GPT-4o may internally use a diffusion head for image decoding.} 
Then, based on empirical observations and analysis of the generated images, we further infer and examine the potential visual encoders, and subsequently propose the complete candidate architectures of GPT-4o (Figure~\ref{fig:complete}).

In addition, our large-scale evaluation reveals \textbf{several limitations in GPT-4o’s generation process}. These include inconsistencies in preserving original content during editing, difficulties in controlling image proportions, automatic cropping effects, high-resolution and over-refinement limitations, challenges in handling complex scenes, occasional color bias, and limitations in generating non-English text. These findings provide valuable insights for future model improvements.



Furthermore, we conduct a comparative analysis of \textbf{multi-round image generation between GPT-4o and Gemini 2.0 Flash\footnote{\url{https://aistudio.google.com/prompts/new_chat}}}, a strong commercial model recently released by Google. Our comparison focuses on four distinct aspects: consistency across multiple edits, instruction comprehension, support for multi-turn editing interactions, and response speed. 
Finally, we also briefly discuss the \textbf{safety and detectability of GPT-4o-generated images}. Interestingly, images generated by GPT-4o are easily detected by state-of-the-art image forensics models. This is likely due to its use of a super-resolution pipeline, which amplifies upsampling interpolation artifacts and makes the images more susceptible to detection by forensic models, or because they contain noticeable watermark-like features.


In summary, our work makes the following key contributions:
\begin{itemize}
    \item We present \textbf{GPT-ImgEval}, the first benchmark to quantitatively and qualitatively evaluate GPT-4o’s image generation capabilities across three well-established benchmarks, including text-to-image generation, editing, and comprehension-guided generation. Our comprehensive results highlight the superior image generation and comprehension capabilities of GPT4o over previous models.
    \item Based on the benchmarking results, we conduct an in-depth analysis that includes: (1) an investigation into the potential underlying architecture of GPT-4o through classifier-based image analysis, and (2) a systematic empirical study of its weaknesses, including common failure modes and generation artifacts.
    \item We further provide a comparative study of multi-round image editing capabilities between GPT-4o and Gemini 2.0 Flash. Additionally, we explore the AIGC safety issue by assessing the detectability of GPT-4o-generated images using existing SOTA image forensic models, revealing that such outputs remain distinguishable due to visible artifacts introduced during upsampling.
\end{itemize}

\section{GPT-ImgEval Evaluation Benchmark}
\label{sec:methods}

\subsection{GPT-4o Image Generation Setup}

\textbf{Dataset.}
In this work, we evaluate GPT-4o’s image generation capabilities using three core datasets: GenEval~\cite{ghosh2023geneval}, Reason-Edit~\cite{huang2024smartedit}, and WISE~\cite{niu2025wise}. Traditional automatic evaluation metrics (such as FID or CLIPScore) primarily measure overall image quality or text-image alignment, but they are not well-suited for fine-grained or instance-level analysis.
\textbf{(1) GenEval} adopts an object-centric framework to assess compositional image attributes, including object co-occurrence, spatial arrangement, counting, and color consistency. This makes it well-suited for evaluating GPT-4o’s ability to control image synthesis based on textual input.
\textbf{(2) Reason-Edit} is a dataset specifically designed for text instruction-based image editing. It covers seven distinct types of editing challenges, testing the model’s capability in spatial understanding, size manipulation, color changes, and commonsense reasoning.
\textbf{(3) WISE} serves as a benchmark for world knowledge-informed semantic evaluation, going beyond simple word-to-pixel mappings. It requires models to generate images grounded in real-world knowledge, including cultural context, temporal and spatial reasoning, and scientific understanding.

\textbf{Automation Scripts.} As of \textbf{April 3, 2025}, GPT-4o does NOT offer an official API for image-generation tasks. To address this limitation, we develop \textbf{custom automation scripts} that interact directly with the GPT-4o web interface. These scripts simulate user input to automatically submit prompts and retrieve the generated images, enabling us to perform large-scale and repeatable evaluations of the model’s image generation capabilities. In order to reduce the interference of the same window context on the model capabilities, the image synthesis corresponding to each prompt is to reopen the window. To support the broader research and developer community, we have open-sourced our automation scripts at \url{https://github.com/PicoTrex/GPT-ImgEval} to facilitate similar workflows for others who wish to conduct automated testing or integrate GPT-4o into their own pipelines.

\subsection{Text-to-Image Generation}


\paragraph{Quantitative Results.}
Results summarized in Table \ref{tab:geneval} evaluate text-to-image (T2I) generation on GenEval~\cite{ghosh2023geneval} across two main model categories: (1) diffusion-based approaches using frozen text encoders for direct prompt-to-image generation, and (2) methods that leverage LLMs or MLLMs to enhance the generation process.
According to the table, \textbf{GPT4o achieves the highest overall score of 0.84, largely outperforming both the frozen text encoder methods and the LLM/MLLM-enhanced approaches}. We observe that ChatGPT-4o demonstrates a clear performance advantage even when compared to the state-of-the-art reasoning-based method GoT~\cite{fang2025got}. It achieves a score of 0.85 on counting tasks, 0.92 on color recognition, 0.75 on spatial localization, and 0.61 on attribute binding. These results highlight the superior capabilities of GPT-4o in spatial reasoning and attribute binding, underscoring its effectiveness in text-guided image generation.

\begin{figure}[!t]
    \centering
    \includegraphics[width=\linewidth]{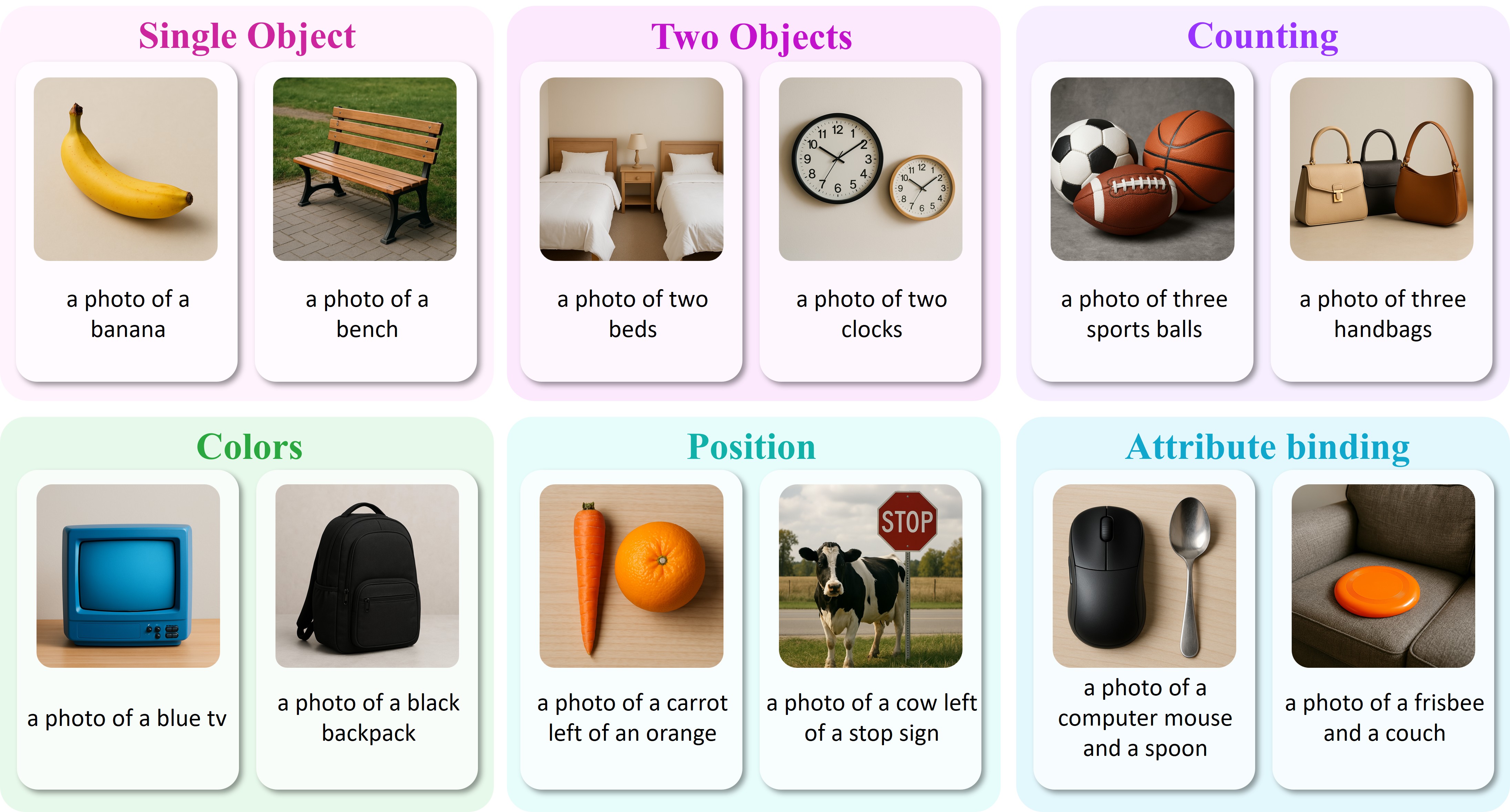}
    \caption{Examples of generation results of GPT4o using GenEval~\cite{ghosh2023geneval}, covering single object, two objects, counting, colors, position, and attribute binding.}
    \label{fig:geneval_case}
\end{figure}

\begin{table*}[!t]
\centering
\caption{Evaluation of text-to-image generation on GenEval~\cite{ghosh2023geneval}. Obj.: Object. Attr.: Attribution.}
\resizebox{\linewidth}{!}{
\begin{tabular}{l|c|c|cccccc}
\toprule
\textbf{Method} & \textbf{Architecture} & \textbf{Overall} & \textbf{Single Obj.} & \textbf{Two Obj.} & \textbf{Counting} & \textbf{Colors} & \textbf{Position} & \textbf{Attr. Binding} \\
\midrule
\multicolumn{9}{l}{\textit{Frozen Text Encoder Mapping Methods}} \\
\midrule
LDM~\cite{rombach2022high} & Diffusion & 0.37 & 0.92 & 0.29 & 0.23 & 0.70 & 0.02 & 0.05 \\
SDv1.5~\cite{rombach2022high} & Diffusion & 0.43 & 0.97 & 0.38 & 0.35 & 0.76 & 0.04 & 0.06 \\
SDv2.1~\cite{rombach2022high} & Diffusion & 0.50 & 0.98 & 0.51 & 0.44 & 0.85 & 0.07 & 0.17 \\
SD-XL~\cite{podell2023sdxl} & Diffusion & 0.55 & 0.98 & 0.74 & 0.39 & 0.85 & 0.15 & 0.23 \\
DALLE-2~\cite{ramesh2022hierarchical} & Diffusion & 0.52 & 0.94 & 0.66 & 0.49 & 0.77 & 0.10 & 0.19 \\
DALLE-3~\cite{ramesh2022hierarchical} & Diffusion & 0.67 & 0.96 & 0.87 & 0.47 & 0.83 & 0.43 & 0.45 \\
SD3 (d=24)~\cite{esser2024scaling} & Diffusion & 0.62 & 0.98 & 0.74 & 0.63 & 0.67 & 0.34 & 0.36 \\
FLUX.1 Dev & Diffusion & 0.66 & 0.98 & 0.81 & 0.74 & 0.79 & 0.22 & 0.45 \\
SD3.5 Large~\cite{esser2024scaling} & Diffusion & 0.71 & 0.98 & 0.89 & 0.73 & 0.83 & 0.34 & 0.47 \\
\midrule
\multicolumn{9}{l}{\textit{LLMs/MLLMs Enhanced Methods}} \\
\midrule
LlamaGen~\cite{sun2024autoregressive} & AR & 0.32 & 0.71 & 0.34 & 0.21 & 0.58 & 0.07 & 0.04 \\
Chameleon~\cite{team2024chameleon} & AR & 0.39 & - & - & - & - & - & - \\
Transfusion~\cite{team2024chameleon} & AR & 0.63 & - & - & - & - & - & - \\
Show-o~\cite{xie2024show} & AR & 0.53 & 0.95 & 0.52 & 0.49 & 0.82 & 0.11 & 0.28 \\
LWM~\cite{liu2024world} & AR & 0.47 & 0.93 & 0.41 & 0.46 & 0.79 & 0.09 & 0.15 \\
SEED-X~\cite{ge2024seedx} & AR & 0.49 & 0.97 & 0.58 & 0.26 & 0.80 & 0.19 & 0.14 \\
Emu3-Gen~\cite{wang2024emu3} & AR & 0.54 & 0.98 & 0.71 & 0.34 & 0.81 & 0.17 & 0.21 \\
TokenFlow-XL~\cite{qu2024tokenflow} & AR & 0.55 & 0.95 & 0.60 & 0.41 & 0.81 & 0.16 & 0.24 \\
Janus~\cite{wu2410janus} & AR & 0.61 & 0.97 & 0.68 & 0.30 & 0.84 & 0.46 & 0.42 \\
JanusFlow~\cite{ma2024janusflow} & AR & 0.63 & 0.97 & 0.59 & 0.45 & 0.83 & 0.53 & 0.42 \\
Janus-Pro~\cite{chen2025janus} & AR & 0.80 & 0.99 & 0.89 & 0.59 & 0.90 & \textbf{0.79} & 0.66 \\
GoT~\cite{fang2025got} & AR & 0.64 & 0.99 & 0.69 & 0.67 & 0.85 & 0.34 & 0.27 \\
HiDream-I1 & AR & 0.83 & \textbf{1.00} & \textbf{0.98} & 0.79 & 0.91 & 0.60 & \textbf{0.72} \\
\midrule
\textbf{GPT-4o} & Unknown & \textbf{0.84} & 0.99 & 0.92 & \textbf{0.85} & \textbf{0.92} & 0.75 & 0.61 \\
\bottomrule
\end{tabular}
}
\label{tab:geneval}
\end{table*}



\paragraph{Qualitative Results.}
Figure~\ref{fig:geneval_case} presents qualitative examples of GPT-4o’s compositional text-to-image generation capabilities across six core evaluation categories in the GenEval benchmark.
In the \textbf{Single Object and Two Objects tasks}, GPT-4o accurately generates clear, well-defined objects corresponding to the prompt (e.g., "a photo of a banana" or "a photo of two clocks"). In \textbf{Counting}, it successfully renders the correct number of items, such as "three sports balls" or "three handbags," demonstrating reliable numerical understanding.

The \textbf{Colors} examples show GPT-4o's ability to associate specific color attributes with the correct objects (e.g., "a photo of a blue TV" and "a photo of a black backpack"), while the \textbf{Position} examples (e.g., "a carrot left of an orange" and "a cow left of a stop sign") highlight its competence in spatial reasoning and object layout.
Finally, the \textbf{Attribute Binding} prompts challenge the model to correctly associate attributes or relationships with multiple objects. GPT-4o handles this effectively, producing well-formed scenes such as "a photo of a computer mouse and a spoon" and "a photo of a frisbee and a couch" without misplacing or merging entities.
These examples collectively demonstrate GPT-4o’s ability to interpret complex compositional prompts and generate coherent, semantically accurate, and visually pleasing images—indicating strong multimodal reasoning and planning capabilities.

\subsection{Image Editing.}


\paragraph{Quantitative Results.}
We also evaluate GPT4o on the image editing task using the Reason-Edit benchmark~\cite{huang2024smartedit}, a benchmark for qualitatively evaluating image editing performance. Following the experimental setup in~\cite{fang2025got}, we employ the GPT Score to evaluate the degree of instruction adherence and the consistency of non-edited regions in image editing tasks. 
\begin{wrapfigure}{l}{0.5\textwidth}
\vspace{-3mm}
  \centering
  \includegraphics[width=\linewidth]{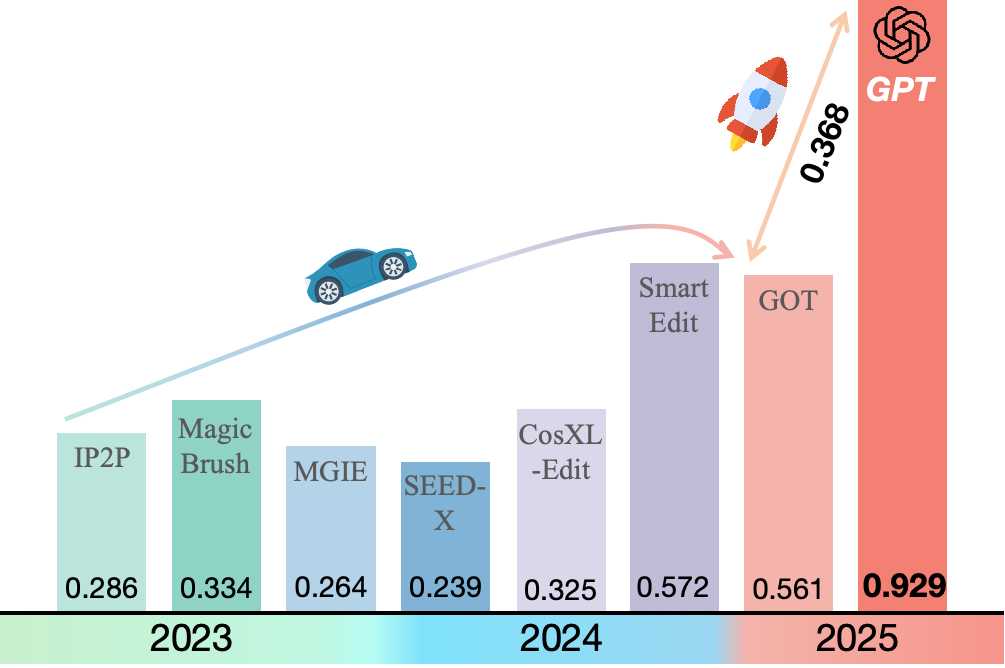}
  \caption{
  Quantitative results of model editing under the Reason-Edit benchmark~\cite{huang2024smartedit}. We compare the performance of GPT4o with seven other SOTA image editing models. We see that GPT4o significantly outperforms other models.
  }
  \label{fig:smart-score}
\vspace{-3mm}
\end{wrapfigure}

As shown in the bar chart (Figure~\ref{fig:smart-score}), \textbf{GPT-4o significantly outperforms all existing image editing methods on the Reason-Edit benchmark, achieving a remarkable score of 0.929.} This represents a substantial leap of +0.357 over the best-performing method prior to 2025 (SmartEdit, 0.572), highlighting the model's powerful instruction-following ability and fine-grained editing control. Compared to state-of-the-art models like GoT (0.561), CosXL-Edit (0.325), and MagicBrush (0.334), GPT-4o sets a new standard in text-guided image editing. GPT-4o demonstrates strong performance in both instruction adherence and image generation quality. The sharp increase in performance demonstrates the potential of integrating large multimodal language models into image editing tasks. Moreover, GPT-4o's image editing process often exhibits inconsistencies in dimensions, color tone, and other global properties. However, such discrepancies are frequently obscured under the GPT-eval Score evaluation framework, which may fail to adequately capture these variations and thus introduce bias in assessing the model’s true performance.

\begin{figure}[t]
    \centering
    \includegraphics[width=\linewidth]{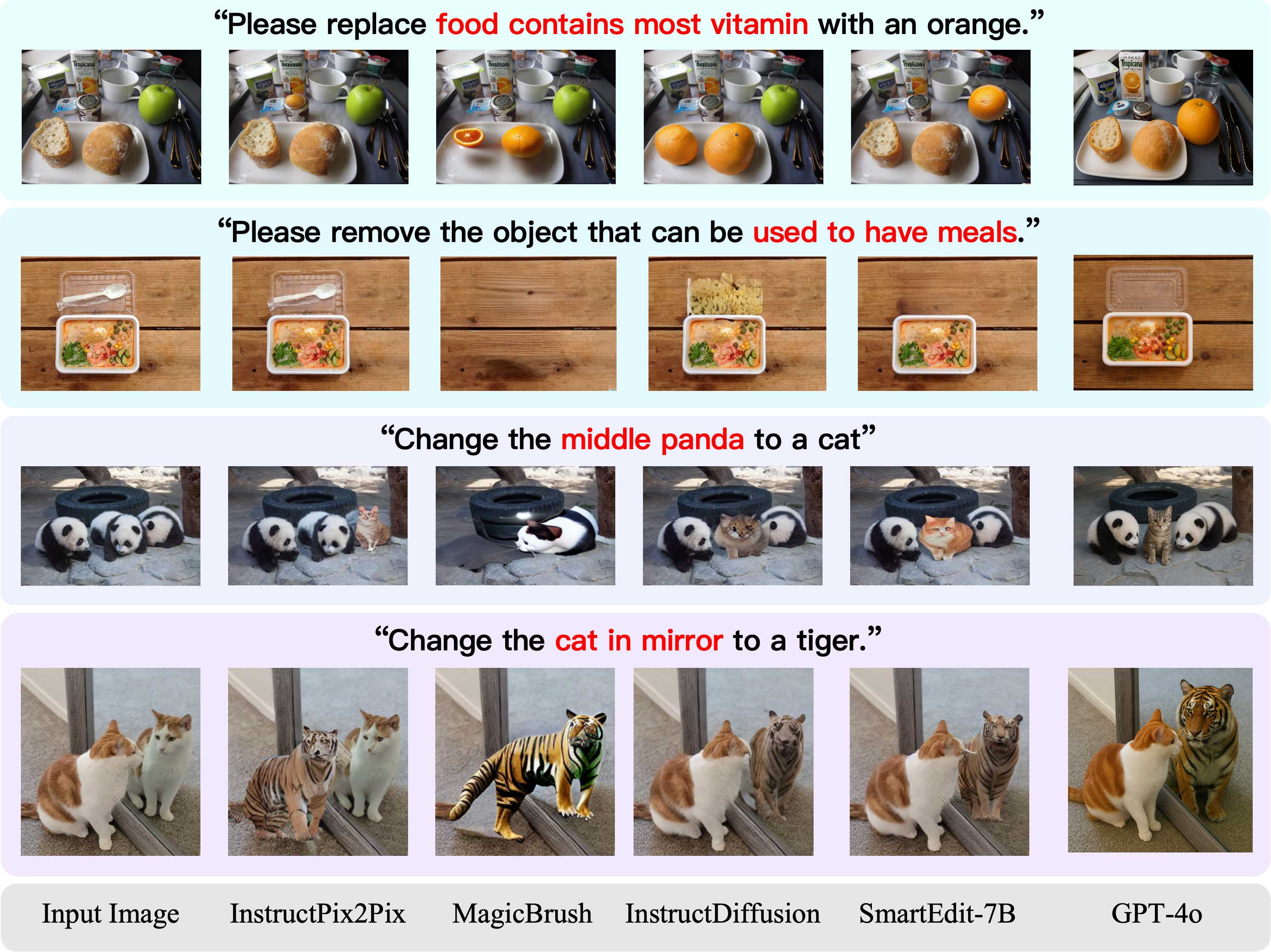}
    \caption{Examples of model editing results. We visualize the qualitative results of GPT4o with the other four SOTA editing generation methods. We use the Reason-Edit~\cite{huang2024smartedit} benchmark for evaluation.}
    \label{fig:smart-case}
\end{figure}

\paragraph{Qualitative Results.}
We present a qualitative comparison of image editing in Figure~\ref{fig:smart-case}, which illustrates the qualitative superiority of GPT-4o across a range of complex image editing instructions. 
For tasks such as object replacement ("replace food contains most vitamin with an orange"), object removal, and attribute-specific substitution ("change the middle panda to a cat"), GPT-4o consistently produces semantically accurate, visually coherent, and contextually aware results. 
Compared to other methods like InstructPix2Pix, MagicBrush, and SmartEdit-7B, GPT-4o shows higher spatial consistency, better localization of edits, and minimal collateral modifications. Moreover, the overall image quality produced by GPT-4o significantly surpasses that of all previous methods. Notably, in the "cat in the mirror" example, only GPT-4o successfully edited the reflection—retaining the real-world background while generating a tiger in the mirror with a matching pose. This task requires fine-grained understanding of semantics and scene structure.


\begin{figure}[t]
    \centering
    \includegraphics[width=\linewidth]{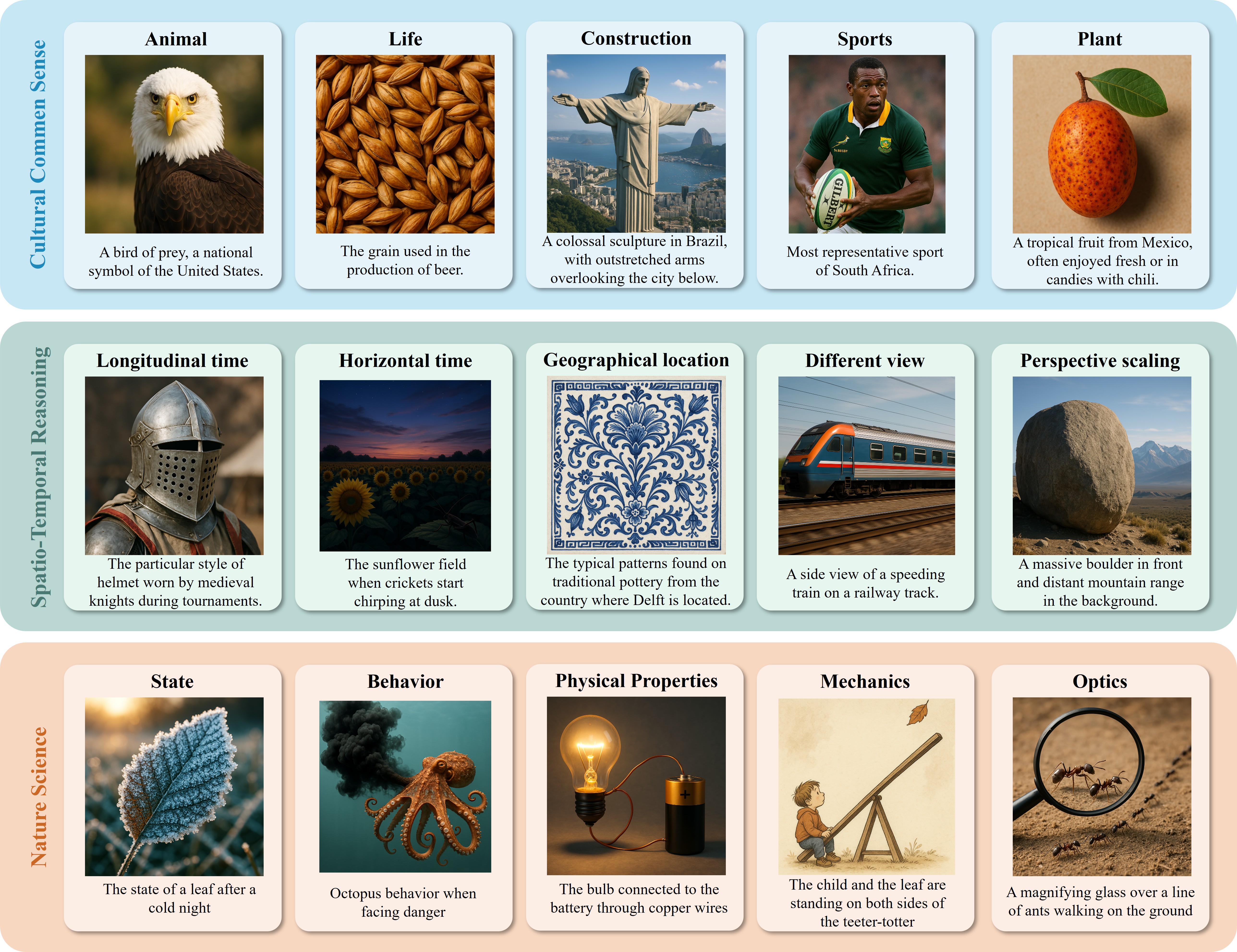}
    \caption{Visual examples of generation results on the WISE benchmark~\cite{niu2025wise}. We visualize the qualitative results of GPT4o under different evaluation scenarios, following the WISE benchmark.}
    \label{fig:WISE}
    \vspace{-2mm}
\end{figure}

\subsection{World knowledge-Informed Semantic Synthesis}

 As existing research and evaluation standards predominantly focus on image realism and shallow text-image alignment, lacking a comprehensive assessment of complex semantic understanding and world knowledge integration in text-to-image generation, in addition to the two above benchmarks, we further evaluate GPT4o on a recent WISE benchmark~\cite{niu2025wise}.  Such tasks require image generation models to possess sufficient world knowledge and reasoning capabilities prior to generation. For instance, given the prompt “Octopus behavior when facing danger,” the model must understand the biological response of an octopus releasing ink. Similarly, the prompt “A colossal sculpture in Brazil, with outstretched arms overlooking the city below” requires the model to recognize and generate the iconic Brazilian landmark—Christ the Redeemer atop Corcovado Mountain.


\begin{table*}[h!]
    \centering
    \caption{Normalized WiScore of different models. }
    \label{tab:WiScore}
    \resizebox{\columnwidth}{!}
    {
        \begin{tabular}{c|c|cccccc|c} 
        \toprule
        Model & Architecture & Cultural  & Time     & Space    & Biology    & Physics    & Chemistry & \textbf{Overall}   \\
        \midrule
        \multicolumn{8}{c}{\textbf{Dedicated T2I}} \\
        \midrule
        FLUX.1-dev~\cite{flux2024} & Diffusion & 0.48  & \textbf{0.58} &\textbf{0.62 } &0.42  &0.51 & \textbf{0.35 }&\textbf{ 0.50} \\
        FLUX.1-schnell~\cite{flux2024} & Diffusion & 0.39  &0.44  &0.50 & 0.31&0.44  &0.26  & 0.40 \\
        PixArt-Alpha~\cite{chen2023pixart} & Diffusion & 0.45  & 0.50& 0.48 & \textbf{ 0.49}&\textbf{0.56} &0.34 &  0.47\\
        playground-v2.5~\cite{li2024playground} & Diffusion &\textbf{0.49 } &0.58  & 0.55&0.43  & 0.48&0.33 & 0.49 \\
        SD-v1-5~\cite{rombach2022high} & Diffusion  & 0.34 & 0.35& 0.32&0.28 &0.29 &0.21 &  0.32\\
        SD-2-1~\cite{rombach2022high} & Diffusion  & 0.30 & 0.38 &0.35 & 0.33 & 0.34&0.21 & 0.32 \\
        SD-XL-base-0.9~\cite{podell2023sdxl} & Diffusion  &0.43  & 0.48 &0.47  &0.44  &0.45 &0.27 & 0.43 \\
        SD-3-medium~\cite{esser2024scaling} & Diffusion  &0.42  & 0.44 &0.48 &0.39  &0.47 &0.29 & 0.42 \\
        SD-3.5-medium~\cite{esser2024scaling} & Diffusion  &0.43  & 0.50 &0.52 &0.41 &0.53 &0.33 & 0.45 \\
        SD-3.5-large~\cite{esser2024scaling} & Diffusion  & 0.44 &0.50 &0.58  & 0.44&0.52 &0.31 & 0.46 \\
        \midrule 
        \multicolumn{8}{c}{\textbf{Unify MLLM}} \\
        \midrule
        Emu3~\cite{wang2024emu3} & AR & 0.34 & 0.45 & 0.48 & 0.41  & 0.45 & 0.27 & 0.39 \\
        Janus-1.3B~\cite{wu2410janus} & AR &0.16 &0.26 &0.35 & 0.28 &0.30 & 0.14&  0.23\\
        JanusFlow-1.3B~\cite{ma2024janusflow} & AR (Diffusion) &0.13 &0.26 &0.28 &  0.20& 0.19&0.11 &  0.18\\
        Janus-Pro-1B \cite{Januspro} & AR & 0.20& 0.28&0.45 & 0.24 & 0.32& 0.16 & 0.26\\
        Janus-Pro-7B \cite{Januspro} & AR & 0.30& 0.37& 0.49 & 0.36 & 0.42 & 0.26 & 0.35 \\
        Orthus-7B-base \cite{Orthus} & AR+Diffusion Head & 0.07 &0.10 &0.12 & 0.15 &0.15 & 0.10&0.10  \\
        Orthus-7B-instruct \cite{Orthus} & AR+Diffusion Head & 0.23 &0.31 &0.38 &0.28  & 0.31&0.20 &  0.27\\
        show-o-demo \cite{show-o} & AR (Diffusion) & 0.28 &0.36  &0.40&  0.23& 0.33& 0.22 & 0.30 \\
        show-o-demo-512 \cite{show-o} & AR (Diffusion) & 0.28 &0.40  &0.48 & 0.30& 0.46 & 0.30 &  0.35\\
        vila-u-7b-256 \cite{vila-u} & AR & 0.26 &0.33  & 0.37 &0.35  &0.39 &0.23 &  0.31\\
        \midrule
        GPT4o & Unknown  & \textbf{0.81}&  \textbf{0.71} & \textbf{0.89} & \textbf{0.83} & \textbf{0.79} & \textbf{0.74} & \textbf{0.80} \\
        \bottomrule
        \end{tabular}
    }
    \vspace{-4mm}
\end{table*}

\paragraph{Quantitative Results.} As results in Table \ref{tab:WiScore} indicate, GPT-4o significantly outperforms existing specialized T2I generation methods and unified MLLM-based approaches in terms of overall WiScore. GPT-4o combines exceptional world knowledge understanding with high-fidelity image generation, demonstrating a dual strength in multimodal generation tasks. This performance gap can be attributed to GPT-4o's strong capacity for world knowledge retention and reasoning, which enables effective integration of knowledge during the image generation process. The results suggest that, within current unified multimodal generation frameworks, the ability to understand and reason about the world does not inherently translate into the capability to visually represent such knowledge with sufficient fidelity and accuracy—yet GPT-4o manages to achieve precisely that.

\paragraph{Qualitative Results.}
We conduct a qualitative comparison in Figure~\ref{fig:WISE}, illustrating GPT-4o's superior performance across several subdomains of World Knowledge-Informed Semantic Synthesis. For example, given the prompt "A bird of prey, a national symbol of the United States", GPT-4o correctly generates a bald eagle. In response to "The particular style of helmet worn by medieval knights during tournaments," it produces an accurate depiction of a fully enclosed medieval helmet with narrow eye slits. For the prompt "The child and the leaf are standing on both sides of the teeter-totter," GPT-4o demonstrates an understanding of weight imbalance by generating a plausibly tilted seesaw. Overall, GPT-4o effectively infers the intended semantics behind the prompts and produces high-quality, semantically aligned images.





\begin{figure}[t]
    \centering
    \includegraphics[width=\linewidth]{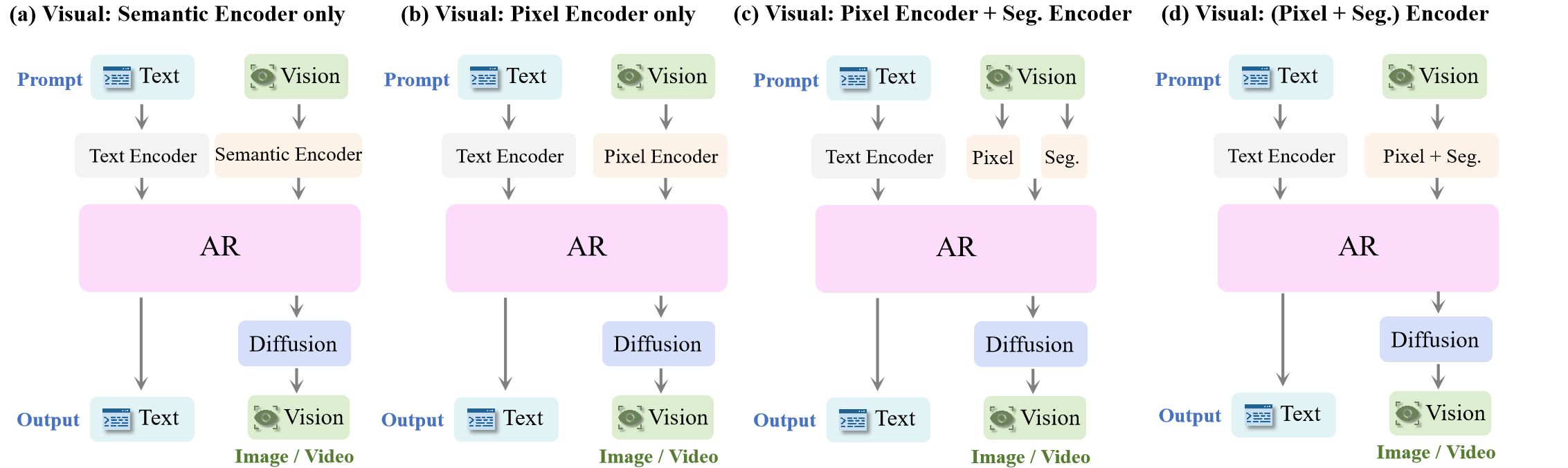}
    \caption{We present a complete architectural speculation, proposing four possible candidates that differ in their choice of visual encoder while all share a diffusion-based head for image decoding.}
    \label{fig:complete}
    \vspace{-4mm}
\end{figure}

\vspace{-2mm}

\section{Potential Architectures Behind GPT4o}

\vspace{-2mm}

We propose three plausible architectural hypotheses that GPT4o might use for image generation (Figure~\ref{fig:archi}).
The three hypotheses are motivated by both the commonly used existing unified architectures~\cite{wang2024emu3, wu2410janus, ma2024janusflow, janus-pro, unitok, show-o, metamorph, llamafusion, transfusion, fan2025unified, song2025dualtoken, chen2025semhitok, nova, seed-x}. 
In the community, the main argument about this topic is the choice of generation head (decoder for image generation), i.e., the choice between architecture-(a) and architecture-(b). 
Below, we first introduce the two architectures in detail, and then provide our analysis and empirical evidence for the discrimination.


\vspace{-2mm}

\paragraph{Hypothesis-1: An VAR-based Architecture with Next-Scale Prediction.}
The first hypothesis posits that GPT-4o employs a VAR-based architecture~\cite{tian2024visual}, as shown in Figure~\ref{fig:archi}(b), which performs next-scale prediction, a progressive refinement strategy that starts with generating a low-resolution, blurry base image and incrementally enhances it to a high-resolution final output. 
This design is inspired by recent VAR generative approaches (e.g., \cite{tian2024visual,han2024infinity}) that scale generation resolution stage by stage.
This hypothesis is reflected by the animation displayed in GPT-4o's image generation interface, where we visually observe the image becoming sharper and more detailed over time.

\vspace{-2mm}

\paragraph{Hypothesis-2: A Hybrid AR Architecture with a Diffusion-Based Head.}
\begin{wrapfigure}{l}{0.4\textwidth}
\vspace{-3mm}
  \centering
  \includegraphics[width=\linewidth]{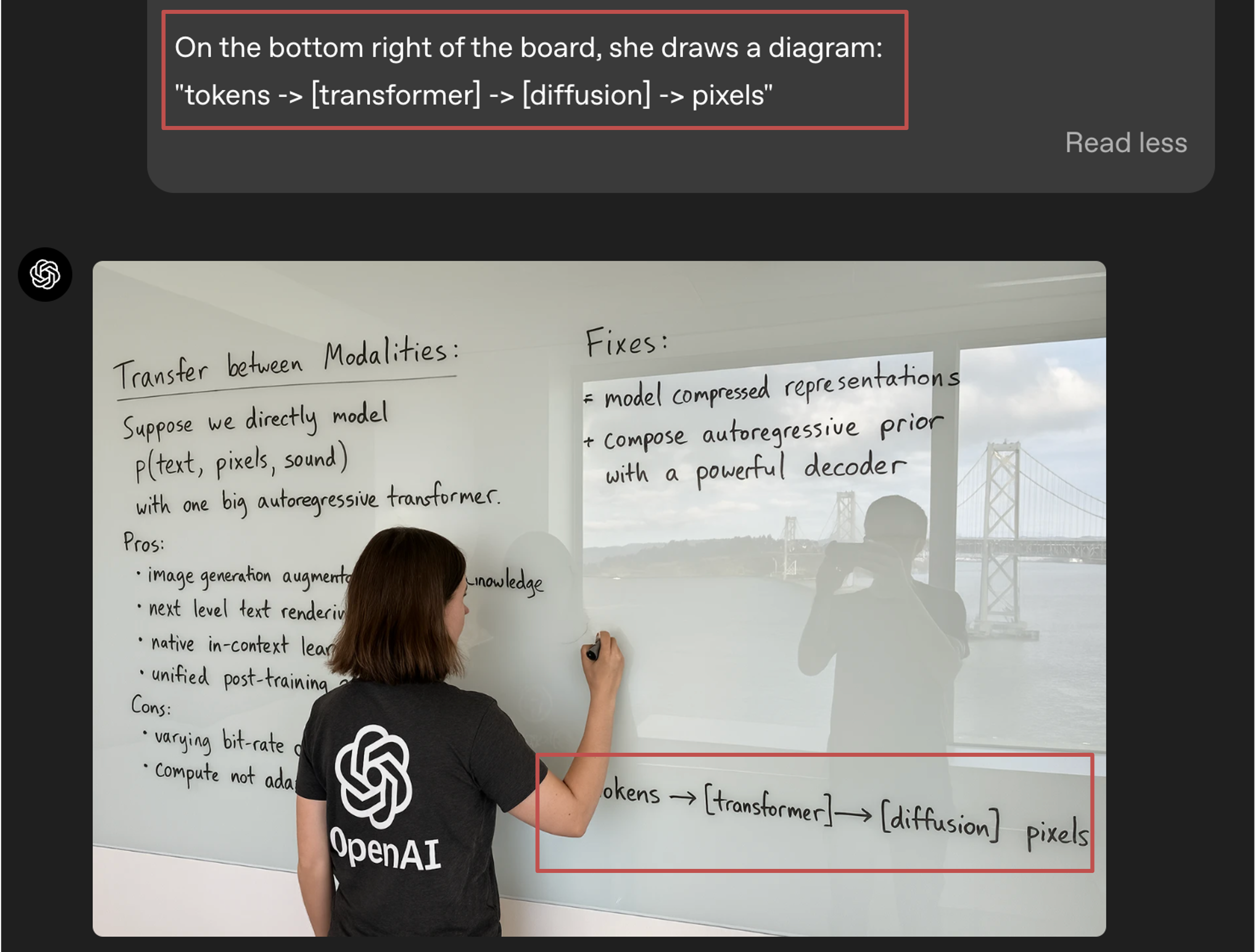}
  \caption{An "easter-egg" example officially provided by OpenAI, which aligns the potential architecture-(a) in Figure~\ref{fig:archi}.}
    \label{fig:oai}
\vspace{-3mm}
\end{wrapfigure}
An alternative hypothesis is that GPT-4o follows a hybrid design, combining a transformer-based AR backbone with a diffusion-based generation head, as illustrated in Figure~\ref{fig:archi}(a). 
In this framework, the AR model first predicts a sequence of intermediate visual tokens or latent representations, which are then used as conditioning input to a diffusion model responsible for decoding the final image, i.e., \textbf{$\text{token} \rightarrow \text{[transformer]} \rightarrow \text{[diffusion]} \rightarrow \text{image pixels}$.}
This hypothesis aligns with descriptions in the OpenAI system card and is consistent with recent efforts to combine the semantic strength of AR models with the visual fidelity of diffusion models.
Interestingly, we identify an "easter-egg" "evidence" provided officially by OpenAI, as shown in Figure~\ref{fig:oai}, which highlights the pipeline of "$\text{token} \rightarrow \text{[transformer]} \rightarrow \text{[diffusion]} \rightarrow \text{image pixels}$" for image generation.

This approach offers a compelling explanation for several behaviors observed in GPT-4o. The model exhibits high image quality, texture diversity, and plausible natural scenes—hallmarks of diffusion-based image generation. At the same time, it shows strong semantic alignment with prompts, suggesting the presence of an AR stage that grounds visual content in language. The hybrid structure also helps explain the "global shift" issues during localized editing, as diffusion models can sometimes struggle to constrain changes to small regions, especially when the conditioning is weak or too coarse.

\begin{figure}[t]
    \centering
    \includegraphics[width=0.95\linewidth]{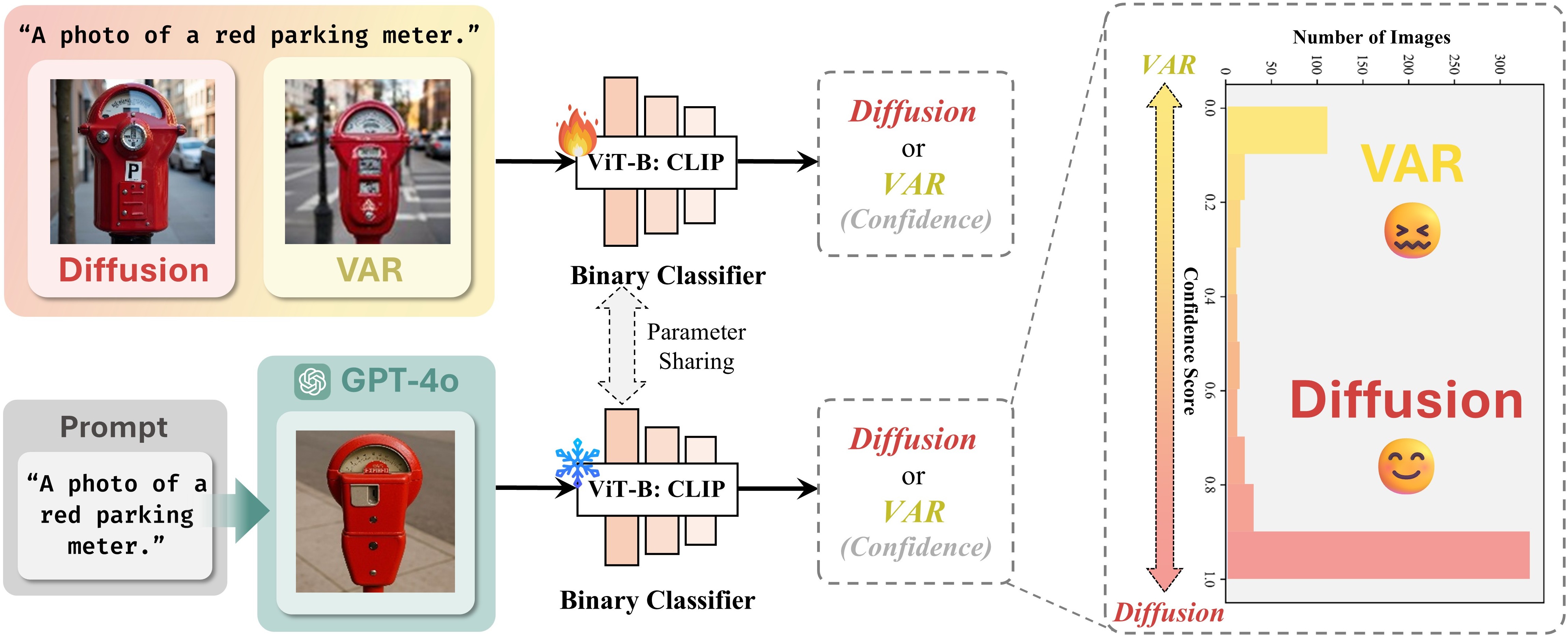}
    \caption{The overall workflow of the proposed model-based discrimination method.}
    \label{fig:discrimin}
\end{figure}

\paragraph{Which Architectures are behind GPT4o?}
To investigate the possible architectures used in GPT4o, we propose a model-based methodology to make the binary discrimination, as illustrated in Figure~\ref{fig:discrimin}. The visual decoder choice is validated based on our empirical experiments. We further infer its potential visual encoder components based on its generated images. Below is the details discussion, respectively.

For the visual \textbf{decoder} component, we conduct a classification-based analysis. We first generate 10,000 images each, using a VAR-based generator and a diffusion-based head, with identical prompts from the GenEval benchmark. Then, a binary classifier is trained to distinguish between the two types of output. When tested on images generated by GPT-4o, the classifier consistently identifies them as diffusion-based. \textbf{This provides strong empirical evidence supporting the hypothesis that GPT-4o employs a diffusion head}, offering new insights into its image-generation mechanism. 

For the visual \textbf{encoder} component, we speculate that \textbf{GPT-4o uses a continuous image tokenizer rather than a discrete one.} This speculation is based on two key observations.
First, GPT-4o cannot perfectly reconstruct an image even when prompted to do nothing; the output images often differ noticeably from the originals in terms of lighting, color, and identity consistency (see Figure \ref{fig:weakness}). 
Since VQ-based discrete tokenizers~\cite{vq} are known for their strong and stable image reconstruction capabilities, this suggests that GPT-4o likely does not use a VQ-based tokenizer.
Second, a recent work, UniTok \cite{unitok}, mentions that applying VQ to images can impair the model's comprehension ability. Therefore, we speculate that GPT-4o likely does not utilize VQ, and instead employs continuous tokens, which is somewhat similar to the approach used in MAR \cite{MAR}, which uses a diffusion loss to model the conditional distribution for generating each token.
We do not have access to the exact architecture of GPT-4o; however, we propose four possible architectures, which are illustrated in Figure \ref{fig:complete}.

\section{Weakness Analysis}

Based on our evaluation results and qualitative inspection of GPT-4o’s outputs, we identified several recurring artifacts that reveal the model’s current limitations in image generation and editing. 
Here, we summarize the key artifact categories observed in GPT-4o's image generation process, highlighting areas where the model struggles to meet expectations for fidelity, consistency, and control. The summary of weaknesses is not limited to our evaluations on the three datasets mentioned above.
In the following, we provide a detailed breakdown of each artifact category.

\paragraph{Inconsistency in image generation.} During image generation, GPT-4o often struggles to perfectly reproduce the input image when no edits are required. Even when prompts explicitly specify "no changes," the model may introduce subtle modifications. This is particularly evident in image dimensions, where the output may exhibit unpredictable aspect ratio changes or automatic edge cropping and rescaling. Such behavior poses significant limitations for applications that demand precise framing or spatial alignment based on the original image dimensions.





\begin{figure}[t]
    \centering
    \includegraphics[width=\linewidth]{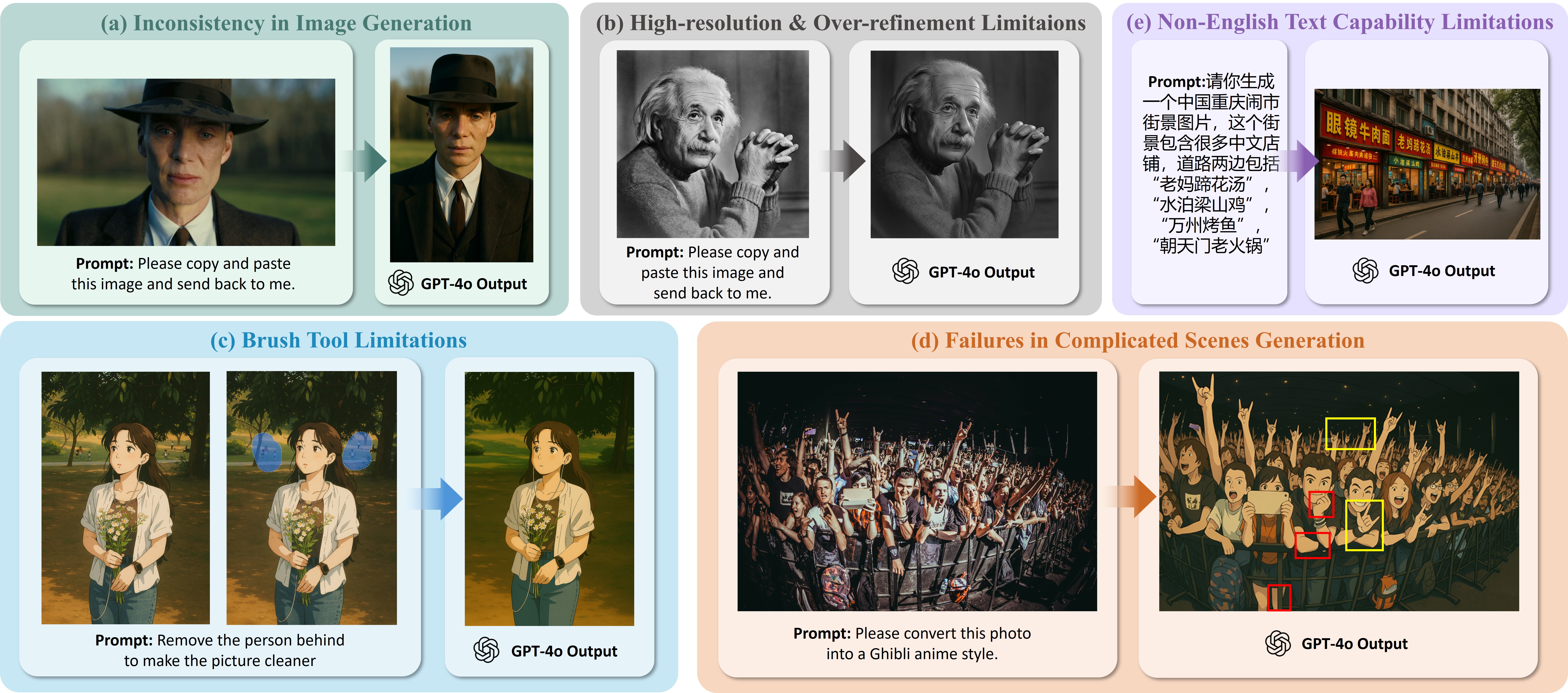}
    \caption{Failure Cases and Limitations of GPT-4o. We identify several scenarios in which GPT-4o may fail, along with common artifacts present in its generated images.}
    \label{fig:weakness}
\end{figure}

\paragraph{High-resolution \& Over-refinement Limitaions.} As illustrated in Figure~\ref{fig:weakness}(b), GPT-4o exhibits a potential bias toward performing super-resolution or image enhancement operations. Even when the prompt explicitly requests a blurry or low-resolution image, the model frequently generates outputs with enhanced clarity and fine detail. This behavior suggests a tendency to prioritize high-frequency visual information, which may stem from internal upsampling modules or biases in the training data. Consequently, GPT-4o struggles to intentionally produce blurred, defocused, or low-detail images, thereby limiting its effectiveness in reproducing certain artistic styles or intended visual effects. Moreover, the model often enriches images with excessive detail—for instance, accurately depicting fine wrinkles on Einstein's face—further reflecting its inherent preference for high-detail synthesis.

\paragraph{Brush Tool Limitations.} Although GPT-4o integrates a brush tool intended for localized image editing, the underlying process still involves regenerating the entire image. As a result, even when only a small region is edited, the output may exhibit unintended changes in global properties such as texture, color, or fine details. In contrast, tools like ComfyUI support true localized inpainting, offering greater stability and control in practical editing applications. Additionally, GPT-4o-generated images frequently exhibit a noticeable warm color bias. In the absence of explicit prompt constraints, the model tends to favor a palette dominated by yellow, orange, and warm lighting. While such outputs may appear visually appealing in certain cases, this bias limits the stylistic diversity of the generated images. The tendency likely stems from imbalanced color distributions in the training data or learned stylistic preferences inherent to large-scale datasets.

\paragraph{Failures in Complicated Scenes Generation.} Although GPT-4o demonstrates remarkable capabilities in generating complex scenes, it still faces significant challenges in producing coherent multi-person scenarios and object-human interactions. As illustrated in Figure~\ref{fig:weakness}(d), the yellow boxes highlight abnormal human poses or anatomical structures, while the red boxes indicate spatially implausible object overlaps. These limitations reflect the model's difficulty in spatial reasoning and maintaining image consistency under high visual complexity.




\paragraph{Non-English Text Capability Limitation.} GPT-4o exhibits superior text generation capabilities, significantly outperforming other models, particularly in rendering English fonts with clarity and consistency. However, its performance in generating Chinese text within complex scenes remains limited. As illustrated in Figure~\ref{fig:weakness}(e), the model frequently produces errors in Chinese signage, such as incorrect fonts or unintended use of traditional characters. This indicates that GPT-4o still faces challenges in non-English text generation. The gap may be attributed to the imbalance between English and Chinese data in training, as well as the inherently greater structural complexity and contextual dependency of Chinese characters.

\section{More Discussion}

\begin{figure}[t]
    \centering
    \includegraphics[width=0.95\linewidth]{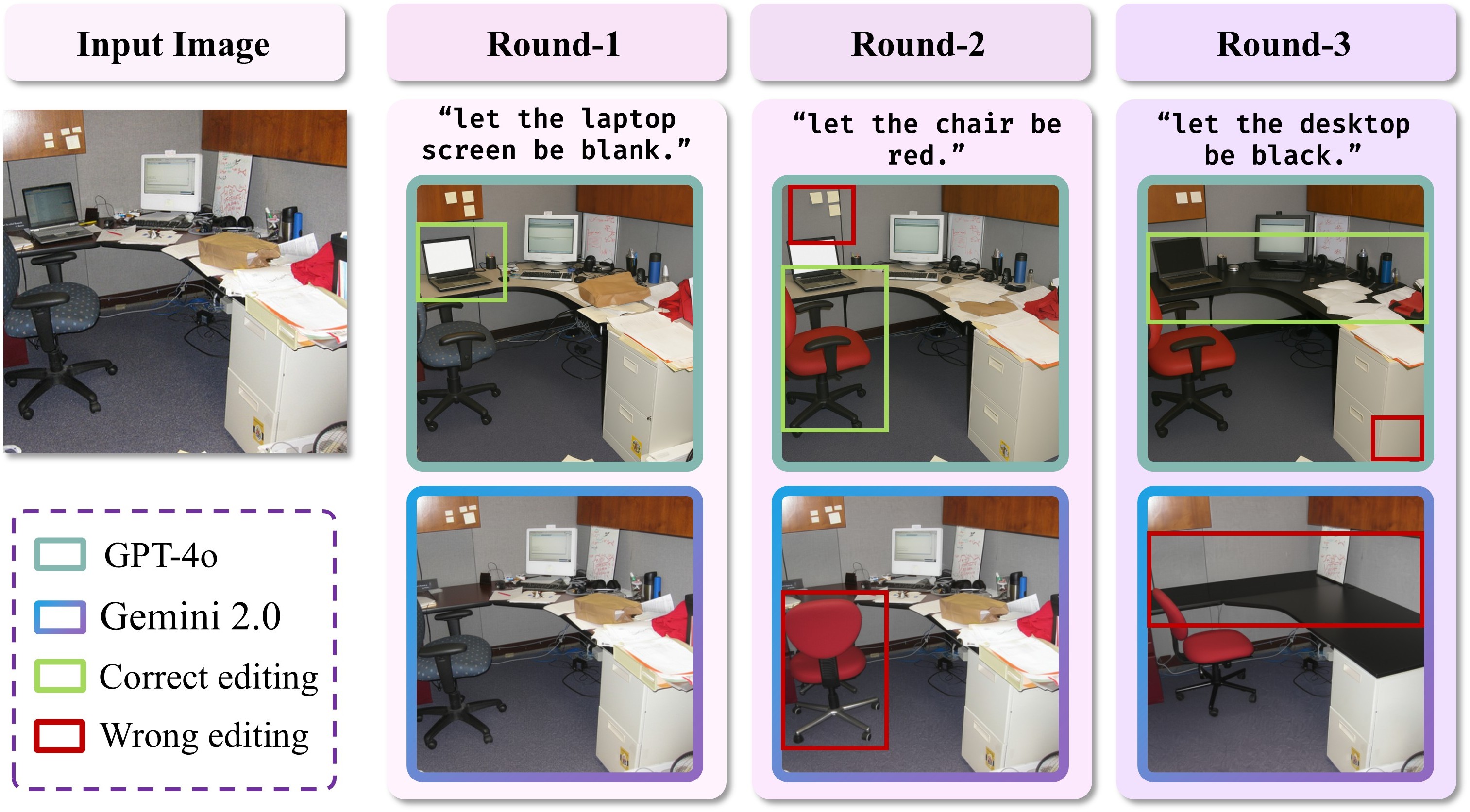}
    \caption{Multi-round generation comparison between GPT-4o and Gemini-2.0 Flash. 
    }
    \label{fig:multi-round}
\end{figure}

\subsection{GPT-4o vs. Gemini 2.0 Flash: Comparative Analysis of Multi-round Generation.}

To compare GPT-4o with another powerful commercial generation model, we conduct a comparative evaluation of GPT-4o and Gemini 2.0 Flash with a focus on image editing consistency, instruction comprehension, multi-turn image editing capabilities, and computational efficiency. Below are our key findings:

\paragraph{Consistency Over Edits.}
Both models exhibit declining consistency with an increasing number of edits. However, GPT-4o performs significantly better than Gemini in this regard. For example, when instructed to change only the color of a chair, GPT-4o correctly alters just the color, while Gemini may unintentionally change the chair’s shape or even its position within the image.

\paragraph{Instruction Comprehension.}
Neither model achieves 100\% accuracy in following instructions. In a test case involving a computer desk, GPT-4o was asked to modify the chair, but instead removed a decorative panel on the wall. Gemini exhibited a more severe failure: not only did it remove the decorative panel, but it also unintentionally erased additional objects in the scene.

\paragraph{Multi-turn Image Editing Dialogues.}
GPT-4o supports multi-turn image editing dialogues, enabling continuous interaction and refinement across multiple image states. In contrast, Gemini 2.0 Flash does not appear to natively support this feature, requiring manual re-uploading of the previous image at each step.

\paragraph{Computational Efficiency.}
In terms of speed, Gemini 2.0 Flash is considerably faster than GPT-4o, making it more suitable for applications requiring rapid responses.

\subsection{For Safety: Is GPT4o-generated Image Detectable?}


GPT-4o demonstrates impressive image-generation capabilities, often producing highly photorealistic results that may appear indistinguishable to the human eye. 
However, our analysis shows that \textbf{these images are still identifiable by current forensic detectors}. 
As shown in Table~\ref{tab:detection}, most existing AI-generated image detectors—including two SOTA models, Effort~\cite{yan2024effort} and FakeVLM~\cite{wen2025spot}—achieve over 95\% accuracy in detecting GPT-4o-generated images. This highlights that, despite their realism, GPT-4o outputs remain within the scope of existing SOTA detection models.

One potential source of detectability lies in \textbf{its internal super-resolution process}. We observe that GPT-4o consistently produces sharp, high-resolution outputs, even when explicitly prompted to replicate blurry or low-resolution images. For instance, when provided with a blurry input image and asked to return it unchanged, GPT-4o instead produces a sharpened, high-resolution version. This suggests a built-in super-resolution mechanism. Supporting this, NPR~\cite{tan2024rethinking}—a forensic model designed to detect upsampling artifacts—achieves 99\% detection accuracy on GPT-4o samples. These findings imply that GPT-4o-generated images may contain distinct, easy-to-recognize artifacts introduced by post-processing steps such as upscaling.

In addition to its technical characteristics, GPT-4o also enforces strong safety safeguards\footnote{\url{https://openai.com/index/introducing-4o-image-generation/}}. The model strictly avoids generating content involving children, recognizable faces, or copyrighted materials such as logos—aligning with OpenAI's robust image safety policies. These restrictions not only enhance user safety but also demonstrate responsible design practices in generative AI deployment.

\begin{table*}[t!]
\centering
\caption{Detection results on GPT4o-generated images. The tested fake images are generated using the prompts in GenEval~\cite{ghosh2023geneval}, and the real ones are from \cite{zeng2025chameleon}. Most detection models are trained on the GenImage dataset~\cite{zhu2023genimage}. $*$ that the training set is from \cite{wen2025spot}.}
\label{tab:detection}
\vspace{1mm}
\resizebox{\textwidth}{!}{
\begin{tabular}{c|ccccccc}
\toprule
Detection Model & CNN-spot~\cite{wang2020cnn} & UnivFD~\cite{ojha2023towards} & CLIP (LoRA) & DRCT~\cite{chendrct} & NPR~\cite{tan2024rethinking} & FakeVLM$*$~\cite{wen2025spot} & Effort~\cite{yan2024effort} \\
\midrule
Accuracy ($\%$) & 73.81 &  75.58 & 77.81 & 88.24 & 78.25 & 99.60 & 94.75 \\
\bottomrule
\end{tabular}%
}
\end{table*}

\section{Conclusion}
\label{sec:conclusion}

This technical report introduces GPT-ImgEval, the first comprehensive benchmark for evaluating GPT-4o's image generation capabilities across three key dimensions: (1) generation quality (via GenEval), (2) instruction-based editing (via Reason-Edit), and (3) comprehension-guided generation (via WISE).
Based on these evaluations, we propose a model-based analysis to infer GPT-4o's underlying architecture and conduct detailed studies to identify its weaknesses and common failure patterns. We further compare GPT-4o with Gemini 2.0 Flash in multi-round image editing and assess the detectability of GPT-4o-generated images.
We hope our work provides meaningful insights and a comprehensive benchmark to inspire future research, promote reproducibility, and drive innovation in image generation and beyond.





{
\small

\bibliographystyle{iclr2017.bst}
\bibliography{references}
}

\clearpage

\section{Appendix}
\label{sec:appendix}

\paragraph{Supplemental Visual Examples of Generation Results.}

\begin{figure}[!h]
    \centering
    \includegraphics[width=\linewidth]{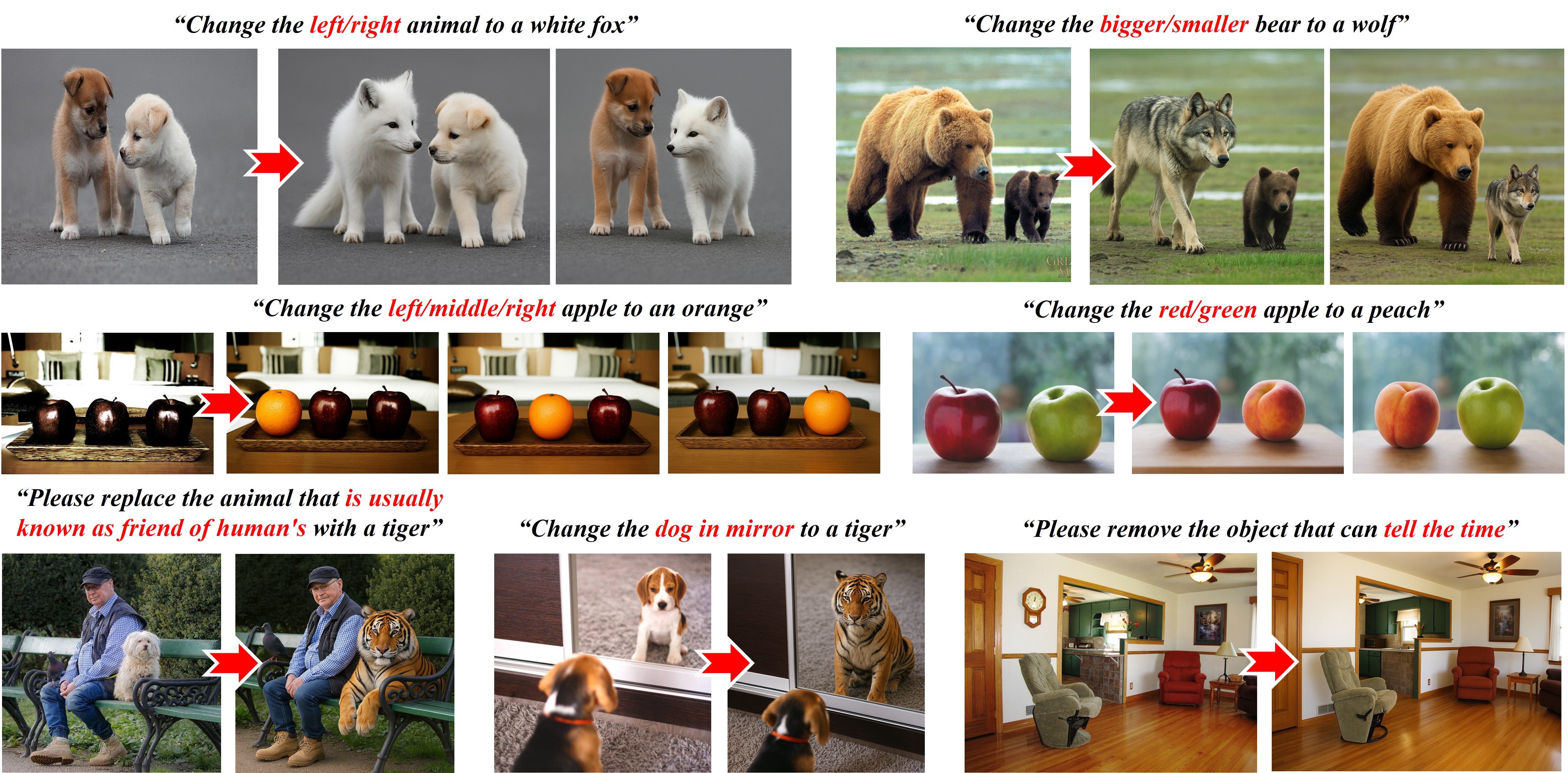}
    \caption{Examples of generation results of GPT4o using Reason-Edit~\cite{huang2024smartedit}.}
    \label{fig:supp_smartedit}
\end{figure}
\begin{figure}[!h]
    \centering
    \includegraphics[width=0.9\linewidth]{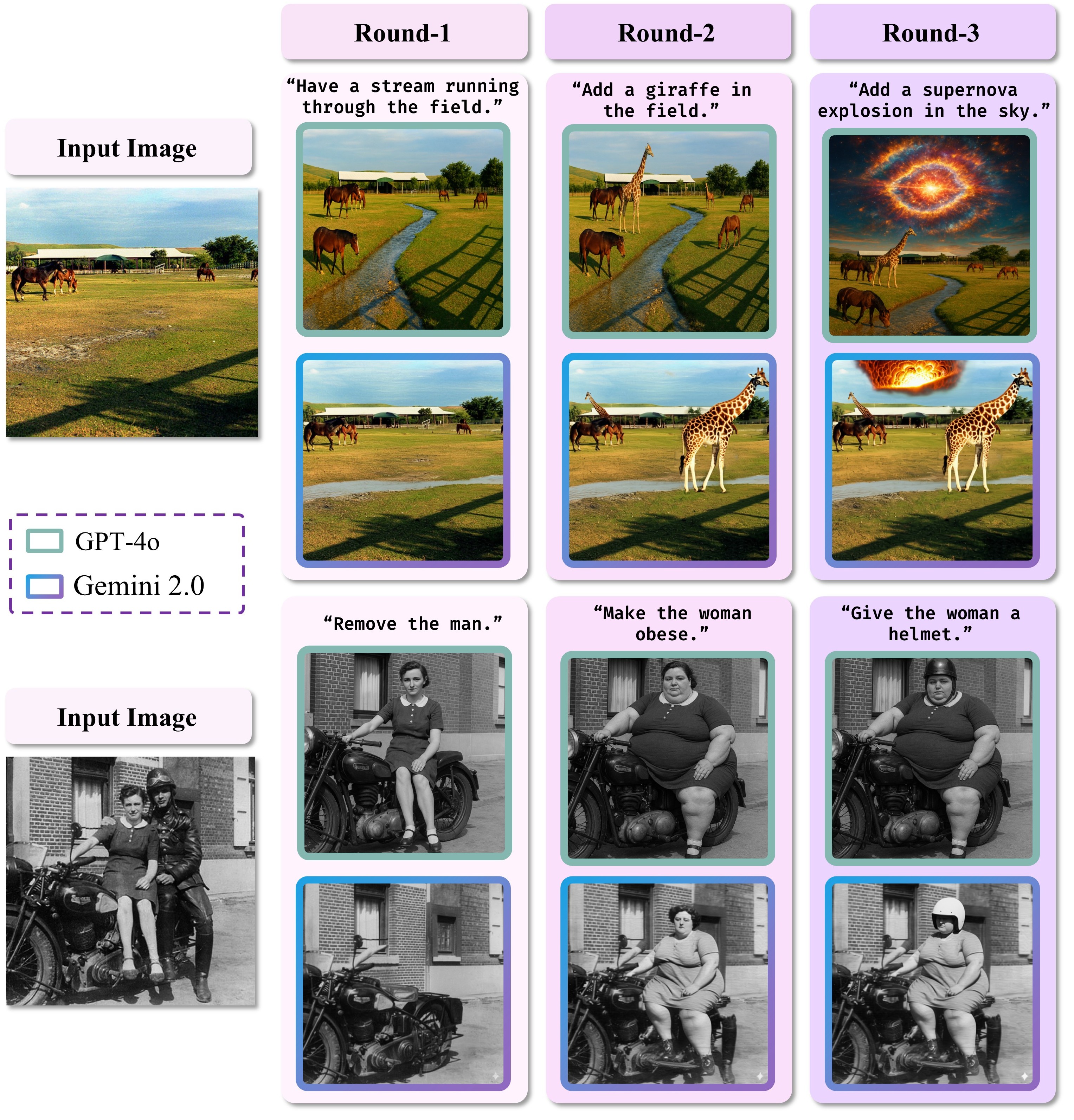}
    \caption{Multi-round generation comparison between GPT-4o and Gemini-2.0 Flash. 
    }
    \label{fig:multi-round2}
\end{figure}

We also include an additional visual example generated by GPT-4o under the Reason-Edit~\cite{huang2024smartedit} benchmark (see Fig.\ref{fig:supp_smartedit}), along with more examples of multi-round editing (see Fig.\ref{fig:multi-round2}).

\paragraph{Details of Model-based Discriminator Between VAR and Diffusion.}
To investigate the potential architecture of GPT-4o, we construct a training dataset and train a discriminator to distinguish between diffusion-based and autoregressive-generated images. The experimental setting is detailed as follows. Flux~\cite{flux2024} and VAR-Infinity~\cite{han2024infinity} are selected to represent diffusion-based and VAR-based image generators, respectively. Using Flux-1 [dev] and Infinity-8B with default settings, we generate images with a resolution of 1024$\times$1024 based on prompts from GenEval~\cite{ghosh2023geneval}, generating 20 images per prompt with different random seeds. A pre-trained CLIP-ViT-Base-16 model is fully fine-tuned over 10 epochs for binary classification using the AdamW optimizer (learning rate: 0.00001, weight decay: 0.0004). GPT-4o-generated images are created from the same prompts from GenEval to construct the test set and mitigate prompt bias.

\end{document}